\documentclass[conference]{IEEEtran}
\IEEEoverridecommandlockouts
\ifCLASSINFOpdf
\else
\fi
\usepackage{graphicx}
\usepackage{float}
\usepackage{comment}
\usepackage{amsmath,amssymb} 
\usepackage{subfigure}
\usepackage{color}
\usepackage{bm}
\usepackage{multirow}
\usepackage{cite}

\makeatletter
\renewcommand{\maketag@@@}[1]{\hbox{\m@th\normalsize\normalfont#1}}%
\makeatother

\begin{document}

\title{Meta-Interpolation: Time-Arbitrary Frame Interpolation via Dual Meta-Learning}

\renewcommand{\thefootnote}{\fnsymbol{footnote}}
\author{\IEEEauthorblockN{Shixing Yu\IEEEauthorrefmark{2}, Yiyang Ma\IEEEauthorrefmark{2}, Wenhan Yang\IEEEauthorrefmark{2}, Wei Xiang\IEEEauthorrefmark{3}, Jiaying Liu\IEEEauthorrefmark{2}\IEEEauthorrefmark{1} \thanks{\IEEEauthorrefmark{1}Corresponding author.\newline \quad This work is supported by the National Key Research and Development Program of China under Grant No. 2018AAA0102702, 
the National Natural Science Foundation of China under Contract No.62172020,  
State Key Laboratory of Media Convergence Production Technology and Systems, and Key Laboratory of Science, Technology and Standard in Press Industry (Key Laboratory of Intelligent Press Media Technology).}}
\IEEEauthorblockA{\IEEEauthorrefmark{2}Wangxuan Institute of Computer Technology, Peking University, Beijing, China}\IEEEauthorrefmark{3}Bigo, Beijing, China}

\maketitle
\pagestyle{plain}
\pagenumbering{arabic}

\begin{abstract}
		Existing video frame interpolation methods can only
		interpolate the frame at a given intermediate time-step, \textit{e.g.} 1/2.
		In this paper, we aim to explore a more generalized kind of video frame interpolation, that at an arbitrary time-step.
		To this end, we consider processing different time-steps with adaptively generated convolutional kernels in a unified way with the help of meta-learning.
		Specifically, we develop a dual meta-learned frame interpolation framework to synthesize intermediate frames with the guidance of context information and optical flow as well as taking the time-step as side information.
		First, a content-aware meta-learned flow refinement module is built to improve the accuracy of the optical flow estimation based on the down-sampled version of the input frames.
		Second, with the refined optical flow and the time-step as the input, a motion-aware meta-learned frame interpolation module generates the convolutional kernels for every pixel used in the convolution operations on the feature map of the coarse warped version of the input frames to generate the predicted frame.
		Extensive qualitative and quantitative evaluations, as well as ablation studies, demonstrate that, via introducing meta-learning in our framework in such a well-designed way, our method not only achieves superior performance to  state-of-the-art frame interpolation approaches but also owns an extended capacity to support the interpolation at an arbitrary time-step.
\end{abstract}


%
\IEEEpeerreviewmaketitle

\section{Introduction}
Video frame interpolation creates non-existent intermediate frames of the input video and maintains the newly generated video to be continuous spatially and temporally and to have a pleasing visual effect.
The technique has been studied widely and becomes a hot research topic in the video processing community.
Its applications range from frame rate up conversion~\cite{up-conversion2,up-conversion1}, novel view synthesis~\cite{deep_stereo}, and inter prediction in video coding~\cite{FrameInterp_Xia,MHT}.
	
Conventional frame interpolation estimates the optical flow of the input frames first, then infer the optical flow at the intermediate time-step, and finally warp the input pixels to the target ones under the guidance of the optical flow~\cite{BakerA,Optical2011}.
This kind of method heavily relies on optical flow estimation, therefore their performance is unstable if there are large motions as the optical flow estimation is usually inaccurate in this case. Furthermore, the conventional optical flow estimation might be time-consuming, therefore the complexity of these methods is usually high.
	
Nowadays, convolutional neural networks (CNN) have been applied to synthesizing the intermediate frames, achieving promising performance in visual quality and time efficiency. All methods can be classified into three branches: 1) direct generation~\cite{direct1} that takes the original frames as input and directly predicts the intermediate frames; 2) flow-guided method ~\cite{liu2017voxelflow,Jiang_2018_CVPR,Niklaus_2018_CVPR,xue2019video} that simulates the align-and-synthesis paradigm; 3) adaptive kernel based method~\cite{Niklaus_2017_CVPR,Bao_2019_CVPR,Niklaus_2017_ICCV} that adopts a flow-free pipeline, where the convolution kernels are learned by passing the original frames through a CNN.
	
All previous methods come across several neglected issues.
1) It is a dilemma whether to adopt the optical flow or not.
Optical flow estimation is an effective representation of motion modeling.
However, its estimation is not robust when large motions are included.
2) Due to the fixed model parameters, all methods can only be applied to handling the interpolation at the given time-step adopted in the training stage, \textit{e.g.} 1/2.
3) As all modules adopt fixed parameters, their adaptivity is not enough to handle different contents and motion conditions accurately and robustly.

Recently, meta-learning, is introduced to increase the model's adaptivity via adjusting the model based on the testing conditions and content of the input images/videos for many computer vision and image processing tasks~\cite{KernelGAN,Shocher_2018_CVPR,Hu_2019_CVPR,fast_adaptation_SR,MAML}.
These works motivate us to address the above-mentioned issues via meta-learning.

In our work, we still follow the paradigm of align-and-synthesis and aim to realize the time-arbitrary video frame interpolation with accurate and robust modeling of motions.
First, to achieve the time-arbitrary video frame interpolation, we build a meta-learned frame interpolation module that takes both the optical flow and time-step as the input for generating the convolutional kernels used in the convolution operations adaptively to produce the predicted intermediate frame.
Besides, it also makes our model more adaptive to different motion contexts and contents when meta-learned adaptive convolutions are introduced.
Second, a meta-learned flow refinement module is introduced to improve the accuracy of the optical flow estimation based on the down-sampled version of the input frames.
As shown in the visual results and the quantitative results, the proposed method outperforms state-of-the-art methods and can provide a better solution for the arbitrary-time video frame interpolation.

\vspace{-6mm}
\begin{figure*}[t]
    	\centering
    	\includegraphics 
        [width=16cm] {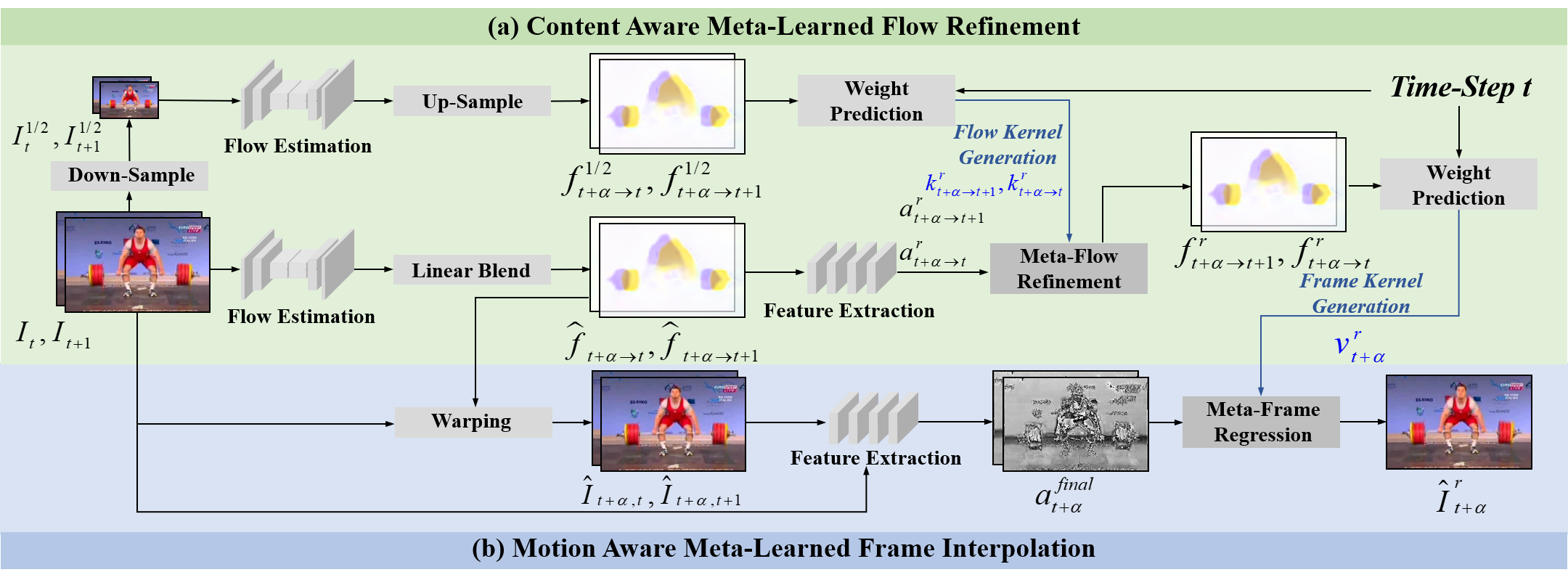}
    	\vspace{-4mm}
    	\caption{
    	The framework of our proposed method.
		(a) The content-aware meta-learned flow refinement module improves the accuracy of the optical flow estimation based on the down-sampled version of the input frames.
		(b) With the refined optical flow and the time-step as the input, the motion-aware meta-learned frame interpolation module generates the pixel-wise convolutional kernels, used to fuse the coarse warped version of the input frames for the frame interpolation.
    	}
    	\vspace{-6mm}
        \label{fig:framework}
\end{figure*}
	
\section{Meta-Interpolation}
\vspace{-1mm}
	\subsection{Motivations}
	
	\vspace{-1mm}
	To address the issues mentioned in introduction, we have the following three motivations in our minds to guide our architecture design:
	\begin{itemize}
		\item \textit{Improving robustness and accuracy of motion estimation}.
		Our architecture inherits the align-and-synthesis paradigm due to the excellent performance this kind of method offers. Furthermore, we hope to inject the motion awareness mechanism into the motion estimation.
		\item \textit{Providing feasibility to decide the time-step of the generated intermediate frame in a flexible way}. It is generally preferred if the framework can take the time-step as the input to support time-arbitrary interpolation under the control of an input time-step.
		\item \textit{Improving adaptivity by making the developed architecture more aware of content and motion contexts.} Thus, in our architecture, we hope to integrate time-step utilization, motion estimation and context perception jointly.
	\end{itemize}
	
	\vspace{-2mm}
	\subsection{Arbitrary-Time Interpolation}
	Conventional video frame interpolation techniques usually take two adjacent frames $I_t$ and $I_{t+1}$ as input and generate the intermediate frame at a given time-step, usually $I_{t+0.5}$.
	The process $\Phi_f(\cdot)$ can be formulated as follows,
	\vspace{-1mm}
	\begin{equation}
	\footnotesize
	\label{eqn:interpolation}
	I_{t+0.5} = \Phi_{fixed}(I_t, I_{t+1}| \theta_f),
	\end{equation}
	\vspace{-6.5mm}

	\noindent where $\theta_f$ is the model parameter and $fixed$ means the fixed time-step. For most of the existing deep-learning video frame interpolation methods, they can only predict the target frame at that time-step.
	
	For the time-arbitrary video frame interpolation problem, the time-step is relaxed as a controllable input, instead of a fixed number, which can be adjusted freely in the testing phase.
	Eqn.~\eqref{eqn:interpolation} can be extended as follows,
	\vspace{-1mm}
	\begin{equation}
	\footnotesize
	\label{eqn:interpolation_arbitrary}
	I_{t+\alpha} = \Phi_{arbitrary}(I_t, I_{t+1}, \alpha | \theta_a),\text{ } 0 < \alpha <1,
	\end{equation} 
	\vspace{-6mm}
	
	\noindent where $\theta_a$ is the model parameter and $arbitrary$ means the arbitrary time-step.
	Many traditional interpolation methods, can be applied to generate the intermediate frame at an arbitrary time-step. 
	In our work, we hope to give deep models such capacity via exploiting the power of meta-learning.
	That is, the model parameters used for interpolation can be dynamically decided in the testing phase based on the input context, \textit{e.g.} frame context, optical flow, and time-step.
	
	
	\vspace{-1mm}
	\subsection{Framework Overview}
	\vspace{0mm}
	The architecture of our framework is briefly shown in Fig.~\ref{fig:framework}.
	In general, the proposed method views video frame interpolation as the result of a convolutional neural network, adaptive instead of fixed based on the testing context.

	Specifically, we develop a dual meta-learned frame interpolation framework to synthesize the intermediate frame $I_{t+\alpha}$ with the guidance of optical flow $f_{t+\alpha,t}$ and $f_{t+\alpha,t+1}$ as well as taking  the time-step $\alpha$ as side information.
	The total pipeline consists of a \textit{content-aware meta-learned flow refinement module} and a \textit{motion-aware meta-learned frame interpolation module}. In the next part, these two modules will be further introduced in detail.
	
	\vspace{-1mm}
	\subsection{Content-Aware Meta-Learned Flow Refinement}	
	\vspace{0mm}
	
	\noindent \textbf{1) Initial Optical Flow Estimation}.
	We first aim to obtain a good motion estimation, fully considering the content information and the time-step.
	To this end, we construct a coarse-to-fine optical flow estimation and refinement pipeline.  
	We first estimate the initial optical flow at both the original resolution and half resolution spaces.
	The flow estimation process is denoted by $E(\cdot)$, then we obtain the optical flow estimations as follows:
	\vspace{-1mm}
	\begin{equation}
	\footnotesize
	\begin{split}
	f_{t \to t+1}^{1/2} = U\left(E \left( I_t^{1/2}, I_{t+1}^{1/2} \right)\right),\\
	f_{t+1 \to t}^{1/2} = U\left(E \left(I_{t+1}^{1/2}, I_{t}^{1/2} \right)\right),
	\end{split}
	\end{equation}
	\vspace{-2mm}
	\begin{equation}
	\footnotesize
	\begin{split}
	f_{t \to t+1} = E \left( I_t, I_{t+1} \right),\\
	f_{t+1 \to t} = E \left( I_{t+1}, I_{t} \right),
	\end{split}
	\end{equation}
	\vspace{-5mm}
	
	\noindent where $U(\cdot)$ is the up-sampling process that projects the frame/flow back to the original resolution space.
	
	\vspace{0mm}
	\noindent \textbf{2) Linear Blend of Flows}.
	Then, based on~\cite{spynet2017}, the flow related to the time-step $\alpha$ can be inferred by the linear blend:
	\vspace{-2mm}
	\begin{equation}
	\footnotesize
	\begin{aligned}
	\hat{f}_{t+\alpha \to t}&=-(1-\alpha)\alpha{{f}_{t \to t+1}}+\alpha^{2}{{f}_{t+1 \to t}},\\
	\label{eqn:linearblend} 
	\hat{f}_{t+\alpha \to t+ 1}&=(1-\alpha)^{2}{{f}_{t \to t+1}}-\alpha(1-\alpha){{f}_{t+1 \to t}}.
	\end{aligned}
	\end{equation} 
	\vspace{-3mm}

	\vspace{0mm}
	\noindent \textbf{3) Flow Kernel Generation}.
	After that, we refine the flow estimation based on these initial estimations.
	In our method, following~\cite{spynet2017}, we seek to utilize the multi-scale information to realize the optical flow information refinement.
	
	The estimated optical flows obtained from $f_{t \to t+1}^{1/2}$ and $f_{t+1 \to t}^{1/2}$ are utilized to provide the context information to infer the refined version of optical flows at the original resolution space.
	Through a cascaded convolutional and fully connected layers, 
	for each location $(i,j)$, input tensors $s_{t+\alpha \to t}^r, s_{t+\alpha \to t+1}^r \in R^{6 \times hw}$ (the superscript $r$ denoting refinement) with both optical flows and time-step information is collected in the following way:
	\vspace{-1mm}
	\begin{equation}
	\footnotesize
	\begin{aligned}
	s_{t+\alpha \to t}^r(i,j)  =
	s_{t+\alpha \to t+1}^r(i,j)  =\\
	(
	f^{1/2}_{t\rightarrow t+1}(i,j), 
	f^{1/2}_{t+1\rightarrow t}(i,j), \alpha, 1-\alpha ).
	\end{aligned}
	\end{equation}
	\vspace{-3.5mm}
	
	After that, the refined side information matrices $s^r_{t+\alpha \to t}$ and $s^r_{t+\alpha \to t+1}$ are calculated
	and feed-forwarded to the given network $K(\cdot)$ to derive the convolutional kernels 
	$k_{t+\alpha \rightarrow t}^r$ and $k_{t+\alpha \rightarrow t+1}^r$ as follows,
	\vspace{-1mm}
	\begin{equation}
	\footnotesize
	k_{t+\alpha \to t}^r = K\left( s^r_{t+\alpha \to t} \right),
	k_{t+\alpha \to t+1}^r = K\left( s^r_{t+\alpha \to t+1} \right).
	\end{equation}

	\vspace{-1.5mm}
	\noindent 	
	\textbf{4) Feature Extraction}. 
	Different from previous methods, in our method, the convolution is not operated directly on the input images.
    Our adaptive convolution operations are applied to the extracted features $a_{t+\alpha \to t}^r$, which are generated via the cascaded convolutional layers $A(\cdot)$ as follows,
	\vspace{-1mm}
	\begin{equation}
	\footnotesize
	a_{t+\alpha \to t}^r = A\left( \hat{f}_{t+\alpha \to t}\right).
	\end{equation}	
	
	\vspace{-1.5mm}
	\noindent 
	\textbf{5) Meta-Flow Refinement}.
	Finally, the refined optical flows are inferred as follows:
	\vspace{-1mm}
	\begin{align}
	\footnotesize
	f^r_{t+\alpha \rightarrow t} = a_{t+\alpha \to t}^r \otimes k_{t+\alpha \to t}^r,
	\end{align}
	\noindent where $\otimes$ denotes the convolutional operation.
	\vspace{-2mm}
	
	\vspace{-0mm}
	\subsection{Motion-Aware Meta-Learned Frame Prediction}
	\vspace{-0mm}
	Based on guidance of refined optical flows, we perform the video frame interpolation.
	
	\vspace{1mm}
	\noindent \textbf{1) Initial Frame Warping}. Based on the initial estimated flows, we can obtain the coarsely warped results as follows,
	
	\vspace{-2mm}
	\begin{equation}
	\begin{aligned}
	\footnotesize
	\hat{I}_{t+\alpha, t} &= W(I_t,f^{r}_{t+\alpha \to t}),\\
	\hat{I}_{t+\alpha, t+1} &= W(I_{t+1},f^{r}_{t+\alpha \to t+1}),
	\end{aligned}
	\end{equation}
	\vspace{-3.5mm}
	
	\noindent where $W(\cdot)$ is backward warping function, which can be
	implemented by bilinear interpolation and is differentiable \cite{Jiang_2018_CVPR}.
	
	\vspace{1mm}
	\noindent \textbf{2) Frame Kernel Generation}.
	We use optical flow information and time stamp as side information to predict the convolutional filters. For each spatial location $(i,j)$, we create the input tensor as follows:
	\vspace{-1mm}
	\begin{equation}
	\footnotesize
	n_{t+\alpha}^r(i,j) = 
	(
	f^{r}_{t+\alpha \to t}(i,j), 
	f^{r}_{t+\alpha\to t+1 }(i,j), \alpha, 1-\alpha ),
	\end{equation}
	\vspace{-5.5mm}
	
	\noindent where $f^{r}_{t+\alpha \rightarrow t}(i,j)$ is the refined optical flow extracted from $I_{t+\alpha}$ to $I_t$ at location $(i,j)$ including two flow fields. The superscripted $f^r$ which denotes the optical flow is a refined version generated from the motion-aware meta-learned optical flow refinement in Sec. 3.4.
	
	After that, the refined side information matrice $n_{t+\alpha }^r$ is feed-forwarded to the given network $V(\cdot)$ to derive the convolutional kernels 
	$v_{t+\alpha}^r$ as follows,
	\vspace{-3mm}
	
	\begin{equation}
	\footnotesize
	v_{t+\alpha }^r = V\left( n^r_{t+\alpha} \right).
	\end{equation} 
	\vspace{-5.5mm}
	
	\noindent
	\textbf{3) Meta-Frame Regression}. 
	To present a coarse-to-fine process, we concatenate the initial warped frames $\hat{I}_{t+\alpha, t}$ and $\hat{I}_{t+\alpha, t+1}$ together with the original input $I_t$, $I_{t+1}$ and extract features through a cascaded convolutional layer $A_{final}(\cdot)$ similar to the feature extraction layer in Sec 3.4 and generate feature maps as follows:
	\begin{equation}
	\footnotesize
	 a_{t+\alpha}^{final} = A_{final}\left(\hat{I}_{t+\alpha, t}, \hat{I}_{t+\alpha, t+1}, I_t, I_{t+1} \right).
	\end{equation}
	\vspace{-5mm}
	
	Finally, we infer the video frame interpolation result as follows:
	\vspace{-1mm}
	\begin{equation}
	\footnotesize
	\hat{I}^r_{t+\alpha } = a_{t+\alpha}^{final} \otimes v_{t+\alpha}^r.
	\end{equation}
	\vspace{-5mm}
	
	\noindent where $\otimes$ denotes the convolutional operation.
	
	\subsection{Implementation Details}
	\noindent \textbf{1) Architecture Details}.
	We choose residual dense network (RDN)~\cite{rdn2018zhang} as the feature extraction network. For frame kernel generation, the network consists of 12 residual dense blocks. Each RDB consists of 8 convolution layers and 64 channels.
	In the flow kernel generation stage, the network's corresponding hyper-parameters are 4, 4, 32.
	The weight prediction networks for both flow refinement and frame interpolation consist of 2 convolution layers, 2 fully connected layers.
	We use the state-of-the-art optical flow prediction algorithm PWC-Net~\cite{pwcnet_2019} for optical flow estimation.
	
	\vspace{0mm}
	\noindent \textbf{2) Loss Function}.
	We denote the ground-truth frame as $I_{t+\alpha}$ and our prediction as $\hat{I}_{t+\alpha}^r$. The reconstruction loss is defined as follows:
	\begin{equation}
	\footnotesize
	L_r = \left \| \hat{I}_{t+\alpha}^r - I_{t+\alpha} \right \|_1.
	\end{equation}
	\vspace{-4mm}
	
	For optical flow refinement, we introduce three kinds of losses: flow loss $L_f$, warping loss $L_w$, and smooth loss $L_s$.
	The flow loss $L_f$ plays a role of regularization for the optical flow refinement by enforcing the consistency between the refined optical flow and the original optical flow (estimated directly by a pretrained model).
	It is denoted as follows:
	\vspace{-1mm}
	\begin{equation}
	\footnotesize
	L_f = \left \| f_{t+\alpha \to t}^r - f_{t+\alpha \to t}^p \right \|_1 + \left \| f_{t+\alpha \to t+1 }^r - f_{t+\alpha \to t+1 }^p \right \|_1.
	\end{equation}
	\vspace{-5mm}
	
	The warping loss $L_w$ regularizes the consistency between the warped frame and adjacent frame, implicitly instructing the refinement of the optical flow. It is represented as follows,
	\vspace{-1.5mm}
	\begin{equation}
	\begin{aligned}
	\footnotesize
	L_w = & \left\| W\left(I_t,f^p_{t+\alpha \to t}\right)- W\left(I_t,f^r_{t+\alpha \to t}\right) \right\|_1  +\\
	\footnotesize
	&\left\|W\left(I_{t+1},f^p_{t+\alpha \to t+1}\right)-
	W\left(I_t,f^r_{t+\alpha \to t+1}\right)
	\right\|_1.
	\end{aligned}
	\end{equation}
	\vspace{-3.5mm}
	
	The commonly used smoothness loss $L_s$ is also adopted to suppress the artifacts in the generated optical flow estimations as follows,
	\begin{equation}
	\footnotesize
	L_s = \left\| \Delta \left(f^r_{t+\alpha \to t}\right) \right\|_1 + \left\| \Delta \left(f^r_{t+\alpha \to t+1}\right) \right\|_1,
	\end{equation}
	\vspace{-4mm}
	
	\noindent where $\Delta(\cdot)$ is the gradient operator.

	In summary, our total training loss is given as follows,
	\vspace{-1mm}
	\begin{equation}
	\footnotesize
	L=\lambda_r L_r+\lambda_f L_f+\lambda_w L_w + \lambda_s L_s ,
	\end{equation}
	\vspace{-6mm}
	
	\noindent where $\lambda_r$, $\lambda_f$, $\lambda_w$ and $\lambda_s$ are the weighting parameters that balance the importance of each term. We empirically set $\lambda_r = 1.0$
	$\lambda_f = 0.02$, $ \lambda_w = 0.2$ and $ \lambda_s = 0.5$.
	
	\vspace{1mm}
	\renewcommand{\baselinestretch}{0.9}
		\begin{table}
    	\begin{center}
    	    \footnotesize
    		\renewcommand{\arraystretch}{1.5}
    		\caption{Quantitative evaluation on the UCF-101, Vimeo90K and Middlebury datasets. The best results are denoted in {\color{red}red} while the second best results are denoted in {\color{blue}blue}.}
    		\vspace{-2mm}		
    		\begin{tabular}{p{1.5cm}cccccccc}
    			
    			\hline\noalign{\smallskip}
    			\multirow{2}{*}{Network} & 
    			\multicolumn{2}{c}{Vimeo90K} & & \multicolumn{2}{c}{UCF-101}&  \multicolumn{1}{c}{MiddleBury}\\
    			& PSNR & SSIM & & PSNR & SSIM &
    			\multicolumn{1}{c}{IE}\\
    			\hline
    			MIND~\cite{MIND_2016_ECCV}  &  33.50 & 0.9429 & & 33.93 & 0.9661 &  \multicolumn{1}{c}{3.35}    \\
    			DVF\cite{liu2017voxelflow}   & 31.54 & 0.9462 & & 34.12 & 0.9631 &  \multicolumn{1}{c}{7.75}     \\
    			SepConv~\cite{Niklaus_2017_ICCV}  &  33.79 & 0.9702 & & 34.69 & 0.9669 &  \multicolumn{1}{c}{2.28}    \\
    			SuperSlomo~\cite{Jiang_2018_CVPR}   & 33.53 & 0.9653 & & 34.26 & 0.9664 &  \multicolumn{1}{c}{2.47}     \\
    			CyclicGen~\cite{liu2019cyclicgen} & 32.10 & 0.9492 & & {\color{red}35.11} & 0.9681 & \multicolumn{1}{c}{2.86} \\
    			CAIN~\cite{cain_2020_aaai} & 34.40 & 0.9715 & & 34.87 & {\color{blue}0.9685} & \multicolumn{1}{c}{2.28} \\
    			DAIN~\cite{Bao_2019_CVPR} & {\color{blue}34.70} & {\color{blue}0.9756} & & {35.00} & 0.9683 & \multicolumn{1}{c}{\color{red}2.04} \\
    			\hline
    			MIN & {\color{red}34.80} & {\color{red}0.9761} & & {\color{blue}35.05} & {\color{red}0.9687} & \multicolumn{1}{c}{\color{blue}2.08} \\
    			\hline
    		\end{tabular}
    		\label{tab:SOTA}
    		\vspace{-5mm}
    	\end{center}
    \end{table}	
    \renewcommand{\baselinestretch}{1.0}
	
    \begin{figure}[t]
		\centering	
		\subfigure{
			\includegraphics[width=1.2cm]{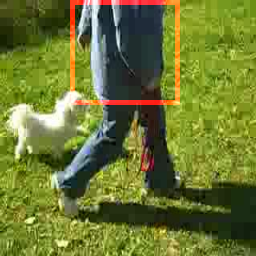}}
		\subfigure{
			\includegraphics[width=1.2cm]{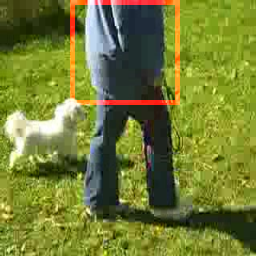}}
		\subfigure{
			\includegraphics[width=1.2cm]{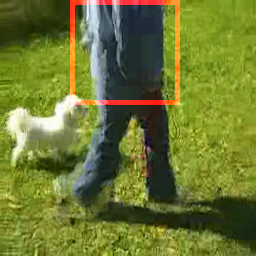}}
		\subfigure{
			\includegraphics[width=1.2cm]{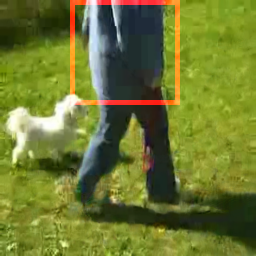}}
		\subfigure{
			\includegraphics[width=1.2cm]{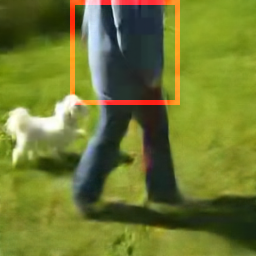}}
		\subfigure{
			\includegraphics[width=1.2cm]{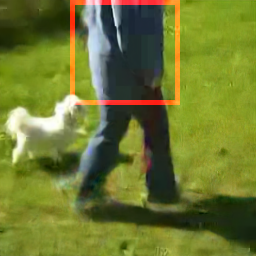}}
		\\ \addtocounter{subfigure}{-6}
		\vspace{-3mm}
		\subfigure[ ]{
			\includegraphics[width=1.2cm]{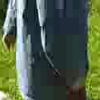}}
		\subfigure[ ]{
			\includegraphics[width=1.2cm]{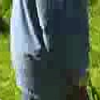}}
		\subfigure[ ]{
			\includegraphics[width=1.2cm]{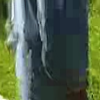}}
		\subfigure[ ]{
			\includegraphics[width=1.2cm]{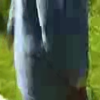}}
		\subfigure[ ]{
			\includegraphics[width=1.2cm]{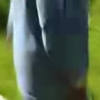}}
		\subfigure[ ]{
			\includegraphics[width=1.2cm]{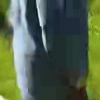}}
		\\
		\vspace{-2mm}
		\caption{
		Visual result of the video frame interpolation by different methods.
		(a) and (b): original inputs $I_0$ and $I_1$. (c)$\to$(e): the results produced by SepConv~\cite{Niklaus_2017_ICCV}, DAIN~\cite{Bao_2019_CVPR}, CAIN~\cite{cain_2020_aaai}, respectively. (f): the result of our method (MIN).
		}
		\vspace{-5mm}
		\label{fig:SOTA0}
	\end{figure}

	\begin{figure}[htbp]
		\centering
		\subfigure[ ]{
			\includegraphics[width=1.2cm]{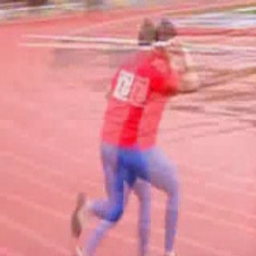}}
		\subfigure[ ]{
			\includegraphics[width=1.2cm]{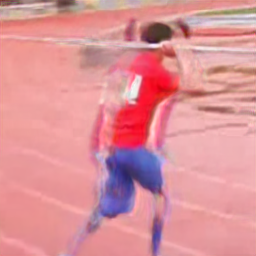}}
		\subfigure[ ]{
			\includegraphics[width=1.2cm]{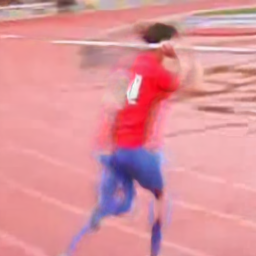}}
		\subfigure[ ]{
			\includegraphics[width=1.2cm]{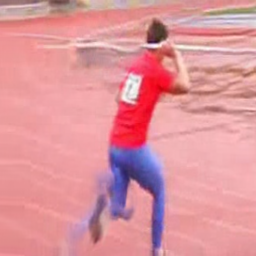}}	
		\subfigure[ ]{
			\includegraphics[width=1.2cm]{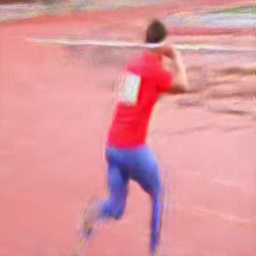}}	
		\subfigure[ ]{
			\includegraphics[width=1.2cm]{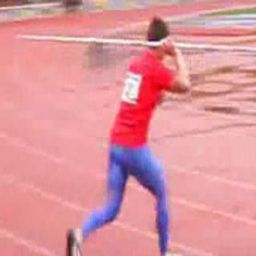}}		
		\vspace{-2mm}
		\caption{
			Visual comparison of the proposed method with different components. 
			(a): Overlapped input. 
			(b): \textbf{MIN-Base}: directly feed-forwarding the input through an RDN. 
			(c): \textbf{MIN-UNR}: MIN-Base + motion-aware meta-learned model.
			(d): SepConv~\cite{Niklaus_2017_ICCV}. 
			(e): \textbf{MIN-Full}: our full version. 
			(f): Ground truth.
		}
		\vspace{-5mm}
		\label{fig:ablation}
	\end{figure}
	
	\renewcommand{\baselinestretch}{0.9}
    \begin{table}[t]
		\begin{center}
    		\footnotesize
    		\caption{Ablation study for \\the motion-aware meta-learned frame prediction and \\the content-aware meta-learned optical flow refinement. \\The best results are denoted in \color{red}{red}.}
    		\renewcommand\arraystretch{1.3}
    		\begin{tabular}{cccc}
    			\hline
    
    			\hline\noalign{\smallskip}
    			Network & Metrics & Vimeo & UCF-101 \\
    			\noalign{\smallskip}
    			\hline
    			\noalign{\smallskip}
    			\multirow{2}{*}{SepConv~\cite{Niklaus_2017_ICCV}}  &  PSNR & 33.79 & 34.69      \\
    			&     SSIM & 0.9702 & 0.9669  \\
    			\hline
    			\multirow{2}{*}{MIN-Base}  &  PSNR & 34.30 & 34.49      \\
    			&     SSIM & 0.9735 & 0.9670  \\
    			\hline
    			\multirow{2}{*}{MIN-UNR}  &  PSNR & 34.47 & 34.61     \\
    			&     SSIM & 0.9746 & 0.9674  \\
    			\multirow{2}{*}{MIN-UNC}  &  PSNR & 34.62 & 34.92     \\
    			&     SSIM & 0.9750 & 0.9682  \\
    		    \hline
    			\multirow{2}{*}{MIN-Full}  &  PSNR & \color{red}{34.80} & \color{red}{35.05}     \\
    			&     SSIM & \color{red}{0.9761} & \color{red}{0.9687}  \\
    			\hline
    			
    			\hline
    		\end{tabular}
    		\label{tab:ablation_motion}
		\end{center}
		\vspace{-5mm}
	\end{table}
	\renewcommand{\baselinestretch}{1.0}
	
    \begin{figure}[t]
		\centering
		\subfigure[$f_{t\rightarrow t+1}$]{
			\includegraphics[width=1.7cm]{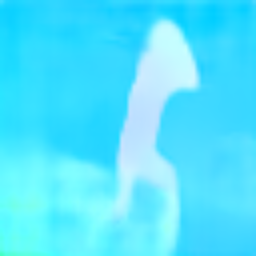}}
		\subfigure[$\hat{f}_{t+\alpha\rightarrow t+1}$]{
			\includegraphics[width=1.7cm]{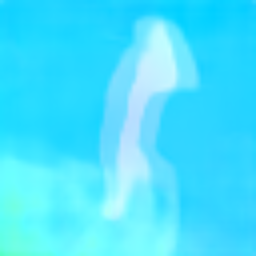}}
		\subfigure[$f^{UNC}_{t+\alpha\rightarrow t+1}$]{
			\includegraphics[width=1.7cm]{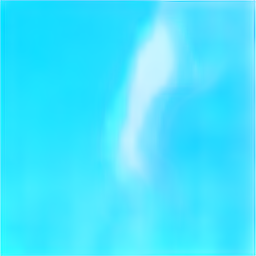}}
		\subfigure[$f^r_{t+\alpha\rightarrow t+1}$]{
			\includegraphics[width=1.7cm]{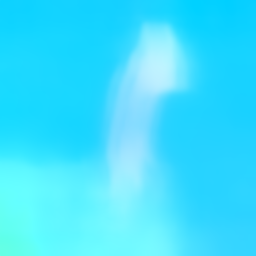}}
		\vspace{-2mm}
		
		\subfigure[$f_{t+1\rightarrow t}$]{
			\includegraphics[width=1.7cm]{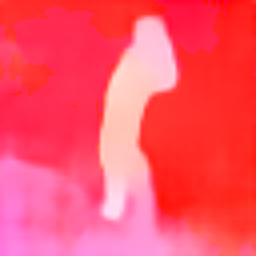}}
		\subfigure[$\hat{f}_{t+\alpha\rightarrow t}$]{
			\includegraphics[width=1.7cm]{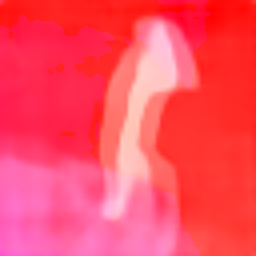}}
		\subfigure[$f^{UNC}_{t+\alpha\rightarrow t}$]{
			\includegraphics[width=1.7cm]{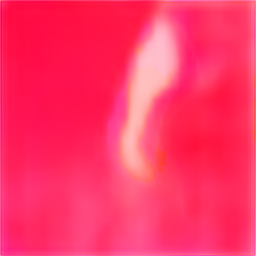}}
		\subfigure[$f^r_{t+\alpha\rightarrow t}$]{
			\includegraphics[width=1.7cm]{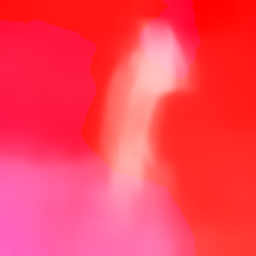}}
		\vspace{-2mm}
		\caption{
			Visualization of the optical flow results of different versions of our method.
		}
		\vspace{-5mm}
		\label{fig:flow_vis}
	\end{figure}

			
	
	\vspace{2mm}
	\section{Experiments}
	\noindent \textbf{1) Datasets and Metrics.}
	We train our proposed video frame interpolation method on Vimeo90K dataset~\cite{xue2019video} and validate on UCF101~\cite{soomro2012ucf101}, Vimeo90K~\cite{xue2019video} and Middlebury benchmark~\cite{BakerA}. Following the setting in \cite{BakerA}, We also report Interpolation Error (IE) on Middlebury benchmark.
	
	\vspace{0mm}
	
	\noindent \textbf{2) Training Strategies and Hyper-Parameter Setting.} We train the network for 20 epochs with a mini-batch size of 2. The initial learning rate is set to 0.001 and a reduce on plateau strategy. Adam\cite{Adam} optimzer is used to update the network parameters, with $\beta_1=0.9,\beta_2=0.999$.

	\vspace{-0mm}
	
	\noindent \textbf{3) Evaluation}.
	We compare the performance of our approach against several SotA methods on quantity and quality. The results are shown in Table~\ref{tab:SOTA} and Fig.~\ref{fig:SOTA0}. Our method is denoted as MIN.
	As we can see in Table ~\ref{tab:SOTA}, our method outperforms almost all the state-of-the-art methods in all metrics.
	
	\vspace{0mm}

	\noindent \textbf{4) Ablation Studies}.
	We analyze the effectiveness of each component and constraint of our method in Table~\ref{tab:ablation_motion}.

	\vspace{-0.5mm}
	\begin{itemize}
	    \item \textit{Motion-Aware Meta-Learned Frame Prediction.} We denote \textbf{MIN-Base} as the model directly feed-forwarding the two concatenated coarse generation results through the RDN. Meanwhile, we also compare the result of another kernel-based method, SepConv~\cite{Niklaus_2017_ICCV}.
	\end{itemize}
	\vspace{-0.5mm}
	
	\vspace{-0.5mm}
	\begin{itemize}
	    \item \textit{Content-Aware Meta-Learned Flow Refinement.} Then, we perform the ablation study on the content-aware meta-learned flow refinement module. We denote the model without the proposed optical flow refinement module as \textbf{MIN-UNR}, the model that only uses the reconstruction loss $L_r$ in the training loss as \textbf{MIN-UNC}.	Besides, we visualize the optical flow to show the effectiveness of this module in Fig.~\ref{fig:flow_vis}. In general, both of our last two versions can well capture the motions of foreground objects.
	\end{itemize}
	\vspace{-0.5mm}
	
	\vspace{-0.5mm}
    \begin{itemize}
	    \item \textit{Visual Comparisons.} We show qualitative results on one image selected from UCF101 in Fig.~\ref{fig:ablation}. Our method successfully generates spatially consistent result in Fig.~\ref{fig:ablation}~(e).
	\end{itemize}
	\vspace{-0.5mm}
	\vspace{-0mm}
	

    \section{Conclusions}
	
	In this paper,  we develop a dual meta-learned frame interpolation framework that is capable to synthesize the intermediate frame at an arbitrary intermediate time-step.
	First, we create a content-aware meta-learned flow refinement module that takes the down-sampled version of the input frames as input to improve the accuracy and robustness of the optical flow estimation.
	Second, a motion-aware meta-learned frame interpolation module generates the convolutional kernels to generate the interpolated frame based on the refined optical flows and the time-step.
	Extensive qualitative and quantitative evaluations demonstrate the superiority of our method and the effectiveness of each component of our method.

	\bibliographystyle{IEEEtran}
	\bibliography{egbib}

\end{document}